\newif\ifanonymous
\anonymousfalse

\newif\ifcameraReady
\cameraReadyfalse

\documentclass{article}

\usepackage{tipa}

\usepackage[normalem]{ulem}
\usepackage{soul}
\usepackage{amssymb,amsmath,mathtools}
\usepackage{amsthm}
\usepackage{stmaryrd}
\usepackage{nicefrac}
\usepackage{algorithmic}
\usepackage{algorithm}
\usepackage{array}
\usepackage{multirow}
\usepackage{scalerel} 
\usepackage{csquotes}
\usepackage[usenames,dvipsnames]{xcolor}
\usepackage{tikz}
\usepackage{pgfplots}
\usepackage{pgfplotstable}
\usepgfplotslibrary{statistics}
\pgfplotsset{
    compat=1.18,
    tick label style={font=\tiny}, 
    label style={font=\small}, 
    legend style={font=\small},
    only if/.style args={entry of #1 is #2}{
        /pgfplots/boxplot/data filter/.code={           \edef\tempa{\thisrow{#1}} 
        \edef\tempb{#2} 
        \ifx\tempa\tempb 
        \else 
             
        \fi 
        } 
    }
}

\definecolor{color1}{HTML}{E29C23}
\definecolor{color2}{HTML}{96AF48}
\definecolor{color3}{HTML}{6580B1}
\definecolor{color4}{HTML}{B53B59}

\usepackage{changepage} 
\usepackage{arydshln} 

\usepackage{makecell}

\usepackage{pifont}
\usetikzlibrary{patterns}
\pgfplotsset{width=1\linewidth,compat=1.7}
\usetikzlibrary{calc,arrows,shapes,decorations,automata,backgrounds,petri,positioning}
\usepackage{orcidlink}
\usepackage[inline,shortlabels]{enumitem}
\usepackage{textcomp}
\usepackage{stfloats}
\usepackage{float}
\usepackage{etoolbox,xspace}

\usepackage{marvosym}
\usepackage{fontawesome}

\usepackage{listings}
\usepackage{url}
\usepackage{verbatim}
\usepackage{booktabs}
\usepackage[most, breakable]{tcolorbox}
\tcbuselibrary{raster}
\usepackage{graphicx}
\usepackage{subcaption}
\usepackage{mathtools} 
\usepackage{arydshln}  
\usepackage{textcomp}
\graphicspath{{./figures/}}
\DeclareGraphicsExtensions{.pdf,.eps,.jpg,.png}
\usepackage{cite}
\usepackage{times}
\usepackage{latexsym}
\usepackage{pifont}

\usepackage{hyperref}
\hypersetup{
	colorlinks=true,
	linkcolor=RedViolet,
	anchorcolor=black,
	citecolor=blue,
	urlcolor=blue,
	breaklinks=true
}

\usepackage{letltxmacro}
\usepackage{xparse}

\makeatletter
\LetLtxMacro\ieeetran@appendix\appendix
\AtBeginDocument{%
	\RenewDocumentCommand{\appendix}{o}{%
		\IfValueTF{#1}{%
			\ieeetran@appendix[#1]%
		}{%
			\ieeetran@appendix%
		}%
	}
}
\makeatother

\usepackage[title,page]{appendix}
\usepackage[capitalise]{cleveref}
\crefname{section}{Sect.}{Sects.}
\crefname{equation}{}{}
\crefname{figure}{Fig.}{Figs.}
\crefname{appendices}{Appx.}{Appx.}
\Crefname{appendix}{Appx.}{Appx.}
\crefname{enumi}{}{}
\crefname{property}{Property}{Properties}
\crefname{lemma}{Lemma}{Lemmas}
\crefname{llemma}{Lemma}{Lemmas}
\creflabelformat{llemma}{#2\thectprop.#1#3}

\usepackage{ijcai25}

\usepackage{soul}
\usepackage{url}
\usepackage[switch]{lineno}

\makeatletter
\renewcommand{\paragraph}{\@startsection{paragraph}{4}{\z@}{-4pt plus -2pt minus -1pt}{-1em}%
  {\normalsize\bfseries\maybe@addperiod}%
}
\newcommand{\maybe@addperiod}[1]{%
  #1\@addpunct{.}%
}
\makeatother

\makeatletter
\renewcommand{\@makefntext}[1]{\noindent\@makefnmark~#1}
\makeatother

\input{macros}

\title{\langname: A Domain-Agnostic Language for Automated Service Regulation\thanks{\langname (\textipa{/"hO:ri:/}) refers to -- in Greek mythology -- the goddesses of order who guarded the gates of Olympus (Homer, \emph{The Iliad}). This paper extends the work-in-progress article~\protect\cite{ICWS/SRG2024}.}
}

\ifanonymous
\author{
    Anonymized Submission \#8853
    \affiliations
    \small{To the IJCAI 2025 Special Track on AI4Tech (Area: \emph{AI4Regulation})}
}
\else
\author{
Yutao Sun$^1$ \and
Mingshuai Chen$^{1{\text{(\Letter)}}}$ 
\and
Tiancheng Zhao$^{2{\text{(\Letter)}}}$ \and
Kangjia Zhao$^1$ \and
He Li$^1$ \and
Jintao Chen$^1$ \and
Zhongyi Wang$^1$ \and
Liqiang Lu$^1$ \and
Xinkui Zhao$^1$ \and
Shuiguang Deng$^1$ \And
Jianwei Yin$^{1{\text{(\Letter)}}}$\\
\affiliations
$^1$Zhejiang University, Hangzhou 310027, China\\
$^2$Binjiang Institute of Zhejiang University, Hangzhou 310053, China\\
\emails
\{m.chen, zjuyjw\}@zju.edu.cn,
tianchez@zju-bj.com
}
\fi

\begin{document}

\maketitle

\setlength{\floatsep}{1\baselineskip}
\setlength{\textfloatsep}{1\baselineskip}
\setlength{\intextsep}{1\baselineskip}

\begin{abstract}
Artificial intelligence is rapidly encroaching on the field of service regulation. However, existing AI-based regulation techniques are often tailored to specific application domains and thus are difficult to generalize in an automated manner. This paper presents 
{\langname}, a unified specification language for modeling (multimodal) regulation rules across a diverse set of domains. We showcase how {\langname} facilitates an intelligent service regulation pipeline by further exploiting a fine-tuned large language model named {\gptname} that automates the {\langname} modeling process, thereby yielding an end-to-end framework for fully automated intelligent service regulation. The feasibility and effectiveness of our framework are demonstrated over a benchmark of various real-world regulation domains. In particular, we show that our open-sourced, fine-tuned {\gptname} with 7B parameters suffices to outperform GPT-3.5 and perform on par with GPT-4o.
\end{abstract}

%
\section{Introduction}\label{sec:introduction}

Service regulation aims to determine whether services are delivered per established norms, rules, and/or standards within a specific context. The rapid advancements in the realm of artificial intelligence (AI) -- particularly breakthroughs in deep neural networks and the swift rise of large language models (LLMs) -- have triggered a recent surge of interest in \emph{intelligent service regulation}. Employing AI in service regulation may substantially improve the degree of automation and accuracy, thereby yielding a significant cost reduction.

\begin{figure}[t]
\centering
\begin{tikzpicture}
\draw (0, 0) node[inner sep=0] {\includegraphics[width=1\linewidth]{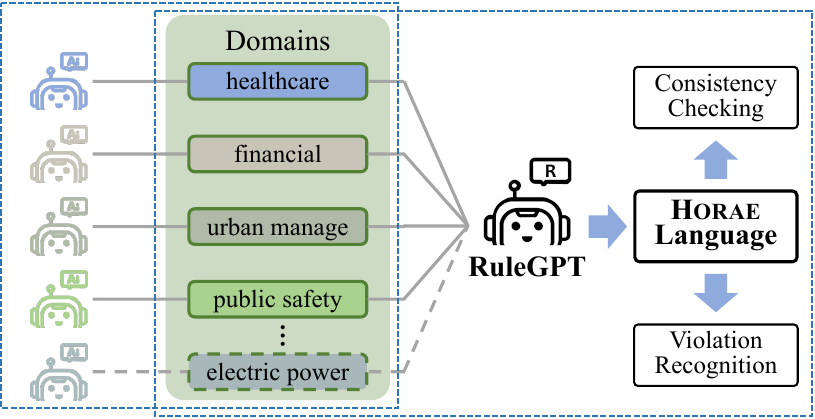}};
\draw (-2.26, -2.5) node {\small (a) plug-and-play methods};
\draw (1.3, -2.5) node {\small (b) {\langname} architecture};
\end{tikzpicture}
\caption{
Conventional plug-and-play methods are often confined to distinct models for specific domains,
thus requiring extensive retraining and resource expenditure. In contrast, {\langname} acts as a unified specification language to model 
regulation rules in a domain-agnostic fashion.}
\label{fig:01-motivation}
\end{figure}

Current AI-based regulation methods predominantly adopt a \emph{plug-and-play} approach: As illustrated in \cref{fig:01-motivation}~(a),
%
regulation industries encompass a wide spectrum of \emph{scenarios} (aka \emph{domains}, e.g., healthcare and financial services).
A common practice is to train a distinct model that caters to a specific
scenario, e.g., models for urban management~\cite{b7}
and e-commerce~\cite{b9}.


The plug-and-play method, however, suffers from two major issues:
\begin{enumerate*}[label=(\roman*)]
	\item \emph{significant resource wastage}: the training and deployment of multiple large-scale AI models tailored for various scenarios
    necessarily incur a model proliferation and thereby substantial computing power consumption and carbon emissions~\cite{b10}; and
	\item \emph{confined adaptability and efficiency}: 
    the procedure of building and training models relies heavily on domain-specific knowledge (e.g., datasets, pre-trained models, and model architectures) of each scenario and is thus difficult to automate for general use.
\end{enumerate*}

In response to these challenges, we propose {{\langname}} -- a unified specification language to model 
regulation rules in a \emph{domain-agnostic} fashion. {\langname} leverages the \emph{zero-shot understanding} capability of LLMs~\cite{DBLP:conf/iclr/WeiBZGYLDDL22} to translate regulation rules from any scenario 
into a structured \emph{intermediate representation} (IR); see \cref{fig:01-motivation}~(b). This representation dissects complex behavior patterns across different domains into a set of fine-grained, readily-detectable events and actions. Consequently, the downstream recognition models and algorithms -- being agnostic to specific domains
-- can utilize a unified rule interface to discharge the regulation tasks.

We show that {\langname} facilitates an intelligent service regulation pipeline by further exploiting a fine-tuned LLM coined {\gptname} to automatically convert regulation rules written in natural languages to the intermediate representation of {\langname}. A formal semantics is further developed for {\langname} to enable rule-consistency checking and quantitative violation recognition (via, e.g., constraint-solving techniques), cf.\ \cref{fig:01-motivation}~(b), thereby yielding an \emph{effective end-to-end framework for fully automated intelligent service regulation}.



\paragraph*{\bf Contributions}
Our main contributions are as follows:
\begin{itemize}
    \item We present {\langname} as a unified specification language to model cross-domain 
    regulation rules. We show that, with a well-designed semantics, {\langname} facilitates core regulation functionalities such as consistency checking and quantitative violation recognition.
    \item We collect a benchmark dataset named {\datasetname} covering a wide range of regulation domains, 
    and thence create a fine-tuned LLM called {\gptname} to automate the modeling process in {\langname}. Both {\datasetname} and {\gptname} are open-sourced to support practical applications in regulation modeling.
    \item We show that {\langname} and {\gptname} admit multimodal rules and enable an end-to-end intelligent service regulation framework. The latter is, to the best of our knowledge, the first framework that admits \emph{fully automated} service regulation with effective \emph{domain unification}.
\end{itemize}
Experimental results demonstrate the feasibility and effectiveness of {\gptname} in automating the modeling process in {\langname} across different real-world regulation domains. In particular, {\gptname} with the size of 7B parameters suffices to outperform GPT-3.5 and perform on par with GPT-4o.


%
\section{General Workflow}\label{sec:workflow}

\begin{figure}[t]
\includegraphics[width=\linewidth]{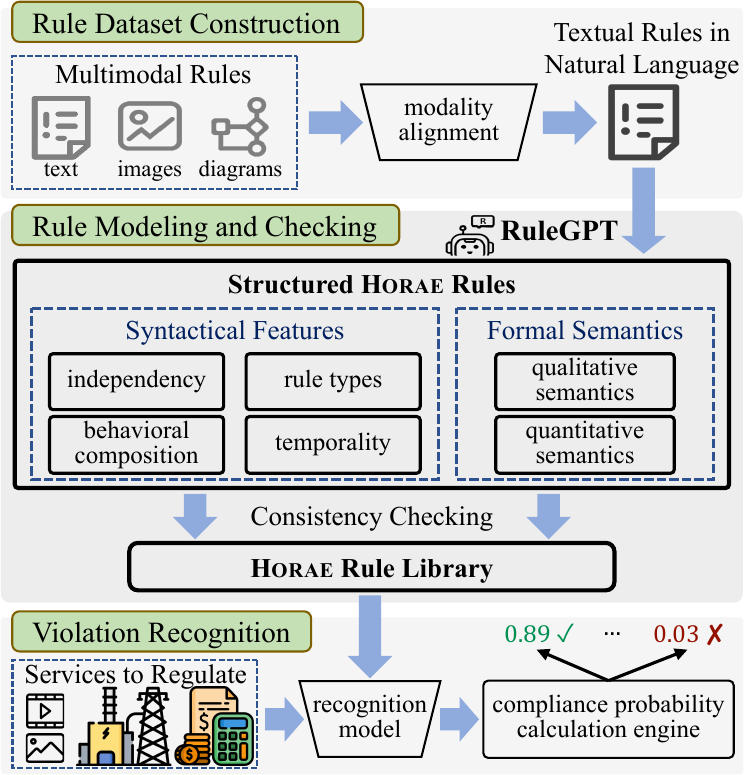}
\caption{{\langname}-steered intelligent service regulation.}
\label{fig:workflow}
\end{figure}

\cref{fig:workflow} sketches an overview of our end-to-end framework of {\langname}-steered intelligent service regulation. This framework consists of the following three major steps:

\begin{enumerate}[label=(\Roman*)]
    \item 
    \emph{Rule Dataset Construction}: This initial step aligns (pre-processed) multimodal regulation rules -- leveraging existing multimodal models -- to the text modality 
    such that rules of different formats can later be interpreted through a \emph{unified medium}, i.e., {\langname} rules.
    \item \emph{Rule Modeling and Checking}: The textual rule dataset is then translated into {\langname} utilizing our fine-tuned {\gptname}. As per the formal semantics of {\langname}, we can check the qualitative and quantitative \emph{consistency} of the rule library to detect potential conflicts before deploying it to downstream regulation tasks.\label{step:rule-modeling-checking}
    \item \emph{Violation Recognition}: The downstream recognition tasks are discharged by multimodal models and algorithms, which assess the \emph{violation probabilities} of basic events in the rule library. These violation probabilities contribute to an overall likelihood of rule violation (computed by a probability calculation engine). 
\end{enumerate}

Our preliminary implementation of the framework indicates that the above steps suffice to produce highly accurate outcomes in a fully automated manner in various real-world domains. This paper focuses on the design principles behind {\langname} and {\gptname} in Step~\ref{step:rule-modeling-checking}. 
The details of aligning multimodal rules to the text modality are provided in \aref{subsec:Multi-Modality} whilst the integration with downstream recognition models and algorithms is subject to future work.


%
\section{Language Design}\label{sec:language}

{\langname} serves as the basis of intelligent service regulation by modeling a set of 
regulation rules in a structured, domain-agnostic fashion. 
We design the syntax and formal semantics as per several key principles, e.g., generality, structuration, automation, and quantification (as detailed in \aref{appx:design-principles}).
These language ingredients constitute the bases of our {\langname} parser (generated by ANTLR 4~\cite{DBLP:conf/oopsla/ParrHF14}); it compiles the text stream of a regulation rule into an abstract tree structure, thereby transforming flat, linear natural language into a structured language with hierarchical patterns.

%

\subsection{Syntax}\label{subsec:syntax}

Our design of the {\langname} syntax follows an \emph{inductive reasoning paradigm}: We first collect a multilingual benchmark set of regulation rules across 50 domains (see details in \aref{appx:benchmark}), then conduct a syntactic analysis over this benchmark set to extract key observations, and finally derive the core patterns and syntax from the body of observations.

Key observations extracted from our benchmark include
\begin{itemize}[leftmargin=16pt,labelwidth=8pt,labelsep=6pt,itemsep=.8mm]
	\item \emph{Independency}: Two textual sentences that are ostensibly disparate in grammatical structure (in terms of their host natural language) may well encode semantically similar regulation rules. For instance, consider the following three rules written in natural languages:
	\vspace*{-1\baselineskip}
	\begin{adjustwidth}{-16pt}{0pt}
		\begin{align*}
			\underbrace{\rulefont{Employees}}_{\mathclap{\textnormal{subject}}}~\underbrace{\rulefont{must}}_{\mathclap{\textnormal{modal verb}}}~\underbrace{\rulefont{wash~hands}}_{\mathclap{\textnormal{verb phrase}}}~\underbrace{\rulefont{before~returning~to~work}}_{\mathclap{\textnormal{prepositional phrase}}}~.
		\end{align*}%
		\vspace*{-3mm}
		\begin{multline*}
			\underbrace{\rulefont{Hand~washing~before~work~resumption}}_{\mathclap{\textnormal{subject}}}~\underbrace{\rulefont{is}}_{\mathclap{\textnormal{linking verb}}} \\[-1mm]
			\underbrace{\rulefont{mandatory}}_{\mathclap{\textnormal{complement}}}~\underbrace{\rulefont{for~all~employees}}_{\mathclap{\textnormal{prepositional phrase}}}~.
		\end{multline*}%
		\vspace*{-5mm}
		\begin{gather*}
			~~~~\includegraphics[width=.96\linewidth]{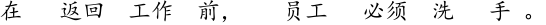}\\[-1mm]
			(\underbrace{\rulefont{when~returning~work~before}}_{\mathclap{\textnormal{temporal adverbial}}},~\underbrace{\rulefont{employees}}_{\mathclap{\textnormal{subject}}}~\underbrace{\rulefont{must}}_{\mathclap{\textnormal{modal verb}}}~\underbrace{\rulefont{wash~hands}}_{\mathclap{\textnormal{verb phrase}}}.)
		\end{gather*}%
	\end{adjustwidth}%
	These three rules (written in English, English, and Chinese, resp.) in fact represent analogous regulation intentions. Hence, the syntax of {\langname} shall be independent of any specific natural language grammar and optimized towards the goal of admitting the most diverse set of 
    intentions with as few grammatical categories as possible.
	\item \emph{Rule Types}: A regulation rule is inherently well-typed, in the sense that, it typically describes certain behavior that is intended to be \emph{enforced}, \emph{recommended}, or \emph{forbidden}:
	\begin{align*}
		&\rulefont{Employees~must~wear~safety~goggles~at~all~times~when~on}\\[-1mm]
		&\rulefont{the~factory~floor.} \tag{\text{enforced}}\\[1mm]
		&\rulefont{It~is~advised~that~all~participants~review~the~safety~manual} \\[-1mm]
		&\rulefont{before~operating~any~machinery.} \tag{\text{recommended}}\\[1mm]
		&\rulefont{No~smoking~is~allowed~within~50~feet~of~the~gas~pumps.} \\[-1mm]
		&\rulefont{\phantom{aa}} \tag{\text{forbidden}}
	\end{align*}%
	{\langname} is thus expected to provide a simple mechanism to specify (a predefined set of) types for regulation rules.
	\item \emph{Behavioral Composition}: The behavioral description of a regulation rule is highly \emph{compositional}, namely, a regulated behavior often appears as a combination of several sub-behaviors via logical connectives, for instance,
    \begingroup
    \allowdisplaybreaks
	\begin{gather*}
		\rulefont{Company~must~conduct~thorough~testing~and~either}\\[-1mm]
		\rulefont{obtain~FDA~approval~or~ensure~compliance~with}\\[-1mm]
		\rulefont{international~health~regulations.}\\[1mm]
		\phantom{\textnormal{decomposition}}~\Downarrow~\textnormal{decomposition}\\[1mm]
		\left(\rulefont{Company~conduct~thorough~testing}\right)~~\wedge\\[-1mm]
		\left(~\left(\rulefont{Company~obtain~FDA~approval}\right)~\vee
		~\left(\rulefont{Company~ensure} \right.\right.\\[-1mm]
		\rulefont{\left.\left.compliance~with~international~health~regulations\right)~\right).}
	\end{gather*}%
    \endgroup%
	Such compositionality is crucial for service regulation as it facilitates the decomposition of a complex regulation problem into a set of sub-problems that can be more easily and accurately solved.
    {\langname} support compositionality by maintaining an abstracted layer of \emph{basic events}, which encode sub-behaviors of a regulated entity and can by logically assembled to describe the entire behavior.
	\item \emph{Temporality}: Temporal properties are yet another important feature in service regulation; they are prominent especially for application domains where timing constraints are crucial, e.g., in financial services:
	\begin{align*}
		&\rulefont{Publicly~traded~companies~have~to~disclose~their~quarterly} \\[-1mm]
		&\rulefont{financial~results~\textit{within~45~days}~by~the~end~of~the~quarter;} \\[-1mm]
		&\rulefont{In~case~any~significant~financial~events~such~as~mergers~or} \\[-1mm]
		&\rulefont{acquisitions~occur~within~these~45~days,\,an~additional~pre}\text{-} \\[-1mm]
		&\rulefont{lim.~report~must~be~submitted~\textit{within~5~days}~of~the~event.}
	\end{align*}
	{\langname} is consequently designed to support temporality by admitting \emph{timestamped events} and \emph{temporal constraints}, which further provide a natural means of modeling regulation rules that are (originally) specified in time-sensitive modalities, e.g., videos; see \aref{subsec:Multi-Modality}.
\end{itemize}


%

Based on these observations, we propose to model a \emph{regulation rule} $R$ in {\langname} per the (abstracted snippet of) syntax:
%
	\begin{align*}
		R \ \  \Coloneqq \ \  & \type~\stmt \TAG{typed rule}\\[1mm]
		\type \ \  \Coloneqq \ \  & \shall \mmid \should \mmid \forbid \TAG{predefined types}\\[1mm]
		\stmt \ \  \Coloneqq \ \  &  \neg \stmt \mmid \stmt \wedge \stmt \mmid 
		\langle \timestamp, \event \rangle \mmid \event \mmid \mathcal{C}(\boldsymbol{\timestamp}) \TAG{statement\footnotemark}\\[-1mm]
		\cdashline{1-3}[2pt/2pt] \\[-4.2mm]
		\event \ \  \Coloneqq \ \  & \obj~\act \mmid \TAG{patterned event}\\
		& \obj~\act~\obj \mmid \obj\mydot\attr~\diamond~\val \mmid \!\!\!\!\\
		& \act~\obj \mmid \act\mydot\attr~\diamond~\val \mmid \cdots
	\end{align*}%
This abstract syntax consists of a \emph{top-level grammar} and a \emph{bottom-level grammar}, as indicated by the dashed line therein. The former combines (possibly timestamped) basic events via logical connectives into a regulation rule of certain type, whilst the latter assembles fine-grained sentence patterns and components into such basic events. Slicing basic events into smaller, detectable ingredients improves the precision of downstream recognition models and algorithms. Below, we provide details of the layered {\langname} syntax.
\footnotetext{$\vee$ and $\rightarrow$ are syntactic sugar expressible by $\neg$ and $\wedge$.}

\paragraph*{\it Top-Level Grammar}
This layer treats \emph{basic events} as the smallest syntactic unit; they will later be interpreted as \emph{propositions} in the formal semantics (see \cref{sec:semantics}). The grammar allows for combining basic events $\event$ via logical connectives and specifying \emph{types} (aka, \emph{execution modes}) of the so-obtained regulation rule -- $\shall$, $\should$, and $\forbid$ for \emph{enforced}, \emph{recommended}, and \emph{forbidden} behaviors, respectively. For rules featuring temporal properties, the corresponding basic event can be associated with a \emph{timestamp} $\timestamp$ signifying its time of occurrence; Moreover, \emph{timing constraints} over timestamps $\boldsymbol{\timestamp} = \{\timestamp_{\textcolor{RedOrange}{1}}, \timestamp_{\textcolor{RedOrange}{2}}, \textcolor{RedOrange}{\ldots}\}$ are collected into $\mathcal{C}(\boldsymbol{\timestamp})$, which acts as a specific form of statement in the rule. 

\paragraph*{\it Bottom-Level Grammar}
This layer describes core patterns of basic events extracted from our rule dataset. Key ingredients include
\begin{enumerate*}[label=(\roman*)]
	\item \emph{action}: the behavior of the basic event;
	\item \emph{object}: actor or recipient of the action -- usually a detectable target;
	\item \emph{attribute}: attributes of objects or actions (selected by the $\mydot$ operator), such as quantity, color, length, etc.; and
	\item $\attr~\diamond~\val$, with $\diamond \in \{<, >, \leq, \geq, =\}$: the \emph{comparison} of some attribute against a given value (e.g., a threshold), which is commonly used in service regulation.
\end{enumerate*}

\subsection{Formal Semantics}\label{sec:semantics}

The formal semantics of {\langname} aims to provide accounts of what a regulation rule adhering to the {\langname} syntax \emph{means} in an unambiguous manner. Such a semantics is essential to represent, interpret, and reason about a typically large set of regulation rules. In particular, it gives a mechanism to check the \emph{consistency} of a rule library in order to detect potential conflicts before deploying it to downstream regulation tasks.

\subsubsection{Qualitative Semantics}\label{subsec:qualitative-semantics} 

We start by formalizing the \emph{qualitative semantics} of {\langname}. Since rule types are fixed in {\langname}, we interpret the (denotational) semantics of a {\langname} rule over its statement. Consider a library of type-free rules:
\begin{align*}
	\rulelib \eeq \left\{ s_1, s_2, \ldots, s_n\right\}~;
\end{align*}%
here, each rule statement $s_k$ with $k = 1, \ldots, n$ is of the form:
\begin{align*}
	s_k \eeq \varphi_k(\boldsymbol{\eventPlain}_k) \wedge \mathcal{C}_k(\boldsymbol{\timestampPlain}_k)~,
\end{align*}%
where $\varphi_k(\boldsymbol{\eventPlain}_k)$ is a \emph{propositional} formula over the set of propositions, i.e., symbolic basic events $\boldsymbol{\eventPlain}_k = \{\eventPlain_{k1}, \eventPlain_{k2}, \ldots\}$ in $s_k$; $\mathcal{C}_k(\boldsymbol{\timestampPlain}_k)$ is the corresponding quantifier-free linear constraints over timestamps\footnote{A timestamp $\timestampPlain_{ki}$ can be absent from $\boldsymbol{\timestampPlain}_k$ if $\eventPlain_{ki}$ is untimed. Assuming linearity of the constraints is necessary to attain decidability (for the qualitative setting) when discharging them via SMT solvers.} $\boldsymbol{\timestampPlain}_k = \{\timestampPlain_{k1}, \timestampPlain_{k2}, \ldots\}$. Without loss of generality, we assume that every rule statement $s_k$ is in \emph{conjunctive normal form} (CNF) over some quantifier-free arithmetic theory $\TT$, i.e., a conjunction of disjunctions of (atomic) arithmetic predicates from $\TT$, for example,
\begin{align}\label{eq:examble-statement}
	s_1 \,=\, (\eventPlain_{11} \vee \eventPlain_{12}) \wedge (\neg \eventPlain_{13} \vee \eventPlain_{14}) \wedge (\timestampPlain_{12} - \timestampPlain_{11} < \timestampPlain_{14})\,.\tag{$\star$}
\end{align}%

Let $\boldsymbol{\eventPlain} \defeq \bigcup_{k=1}^{n} \boldsymbol{\eventPlain}_k$ and $\boldsymbol{\timestampPlain} \defeq \bigcup_{k=1}^{n} \boldsymbol{\timestampPlain}_k$ be, respectively, the set of all basic events and timestamps in $\rulelib$.
A \emph{qualitative interpretation} of $\rulelib$ is a (total) mapping:
\begin{align*}
	\interp \colon\ \ \boldsymbol{\eventPlain} \uuplus \boldsymbol{\timestampPlain} \tto \BB \uuplus \PosReals~,
\end{align*}%
where $\uplus$ denotes disjoint union; $\interp$ thus interprets every basic event over the Boolean domain $\BB \defeq \{\TRUE, \FALSE\}$ and every timestamp over the set of non-negative real numbers $\PosReals$. Let $\interps$ be the set of all possible qualitative interpretations.

We define the \emph{qualitative semantics} of $\rulelib$ as 
\begin{align*}
	\semQL{\rulelib}\colon\ \ \interps \tto \BB, \qquad \interp \mmapsto \bigwedge\nolimits_{k=1}^{n} s_k(\interp)~,
\end{align*}%
where $s_k(\interp)$ denotes the substitution of interpretation $\interp$ in $s_k$. The qualitative semantics of rule statement $s_k$, i.e., $\semQL{s_k}$, is then a projection of $\semQL{\rulelib}$ over $\boldsymbol{\eventPlain}_k$ and $\boldsymbol{\timestampPlain}_k$.
We say that the rule library $\rulelib$ is \emph{qualitatively consistent} if there exists an interpretation under which $\semQL{\rulelib}$ evaluates to $\TRUE$, i.e.,
\begin{align}\label{eq:qual-consistency}
	\exists \interp \in \interps \colon\ \ \semQL{\rulelib}(\interp) \eeq \TRUE~.\tag{$\dagger$}
\end{align}%
The qualitative consistency of $\rulelib$ as per \cref{eq:qual-consistency} can be decided (over the quantifier-free mixed linear integer and real arithmetic~\cite{DBLP:conf/fmcad/0001BT14}) by various off-the-shelf satisfiability modulo theories (SMT) solvers, e.g., Z3~\cite{z3} and \textsc{cvc5}~\cite{DBLP:conf/tacas/BarbosaBBKLMMMN22}.

\subsubsection{Quantitative Semantics}\label{subsec:quantitative-semantics} 

The proposed qualitative semantics $\semQL{\rulelib}$ does not address the \emph{quantitative} aspects of rule satisfaction, i.e., the likelihood of it being satisfied. Such quantitative aspects are crucial for intelligent service regulation since the underlying recognition models and algorithms inherently produce imprecise results (measured by certain confidence factors). We thus extend the qualitative semantics to characterize quantitative satisfaction.

Let $\prob \defeq [0, 1] \cap \Reals$ be the domain of probabilities. Given a rule library $\rulelib$, the \emph{quantitative interpretation} of $\rulelib$ is
\begin{align*}
	\interpQT \colon\ \ \boldsymbol{\eventPlain} \uuplus \boldsymbol{\timestampPlain} \tto \prob \uuplus \PosReals~,
\end{align*}%
i.e., it interprets every basic event $\eventPlain_{ki}$ as the \emph{probability} $p(\eventPlain_{ki}) \in \prob$ of it being $\TRUE$ (cf.\ $\BB$ for the qualitative case). Let $\interpsQT$ be the set of all possible quantitative interpretations.

Similarly, we define the \emph{quantitative semantics} of $\rulelib$ as 
\begin{align*}
	\semQT{\rulelib}\colon\ \ \interpsQT \tto \prob, \qquad \interpQT \mmapsto \prod\nolimits_{k=1}^{n} \measure(s_k(\interpQT))~,
\end{align*}%
where $\measure(s_k(\interpQT))$ denotes the probability that $s_k$ is satisfied under $\interpQT$, which can be computed recursively as
\begin{align*}
	\measure(s_k(\interpQT)) \!= \!\left\{\!\!\!
	\begin{array}{l@{\hspace{-3.6cm}}r}
		1, \quad &\text{if $s_k(\interpQT)$ is logically equivalent to $\TRUE$}\\
		0, \quad &\text{if $s_k(\interpQT)$ is logically equivalent to $\FALSE$}\\
		p(\eventPlain_{ki}), \quad &\text{if $s_k = \eventPlain_{ki}$}\\
		1- \measure(s(\interpQT)), \quad &\text{if $s_k = \neg s$}\\
		\measure(s(\interpQT)) \cdot \measure(s'(\interpQT)), \quad &\text{if $s_k = s \wedge s'$}\\
		1 - \measure(\neg s(\interpQT)) \cdot \measure(\neg s'(\interpQT)), \quad &\text{if $s_k = s \vee s'$}\\
	\end{array}
	\right.
\end{align*}%
Analogously, the quantitative semantics of rule statement $s_k$, i.e., $\semQT{s_k}$, is then a projection of $\semQT{\rulelib}$\! over $\boldsymbol{\eventPlain}_k$ and $\boldsymbol{\timestampPlain}_k$.
For instance, given the quantitative interpretation:
\begin{align*}
	\interpQT \colon\quad\eventPlain_{11} &\mapsto 1, &\eventPlain_{12} &\mapsto 0, &\eventPlain_{13} &\mapsto \nicefrac{1}{2}, &\eventPlain_{14} &\mapsto \nicefrac{1}{3}, \\
	\timestampPlain_{11} &\mapsto 3.5, &\timestampPlain_{12} &\mapsto 6, &\timestampPlain_{13} &\mapsto 11, &\timestampPlain_{14} &\mapsto 3~,
\end{align*}%
The quantitative semantics of the statement $s_1$ in \cref{eq:examble-statement} is
\begin{align*}
	\semQT{s_1}\!(\interpQT) \eeq (1 - 0 \cdot 1) \cdot (1 - \nicefrac{1}{2} \cdot \nicefrac{2}{3}) \cdot 1 \eeq \nicefrac{2}{3}~.
\end{align*}%

We say that the rule library $\rulelib$ is \emph{quantitatively consistent} if there exists a quantitative interpretation under which $\semQT{\rulelib}$\! exhibits a positive satisfaction probability, i.e.,
\begin{align}\label{eq:quat-consistency}
	\exists \interpQT \in \interpsQT \colon\ \ \semQT{\rulelib}\!(\interpQT) \GG 0~.\tag{$\ddagger$}
\end{align}%
The quantitative consistency of $\rulelib$ as per \cref{eq:quat-consistency} can be decided (over the non-linear real arithmetic~\cite{Tarski51}) by dedicated SMT solvers, e.g., dReal~\cite{DBLP:conf/cade/GaoKC13} and SMT-RAT~\cite{DBLP:conf/sat/CorziliusKJSA15}.

\begin{remark}
\emph{Event correlation} remains as a challenge in consistency checking: 
Basic events $\boldsymbol{\eventPlain}$ from the same rule library $\rulelib$ may well be \emph{semantically correlated} (in terms of natural language) with each other, especially for events across different rule statements. We address this problem through \emph{event abstraction}, i.e., abstracting these events written in natural languages into a set of symbolic propositions while preserving semantic correlations; see details in \aref{appx:event-abstraction}.
\qedT
\end{remark}

\section{Automation}\label{sec:automation}

\begin{figure*}[t]
	\centering
	\includegraphics[width=.95\linewidth]{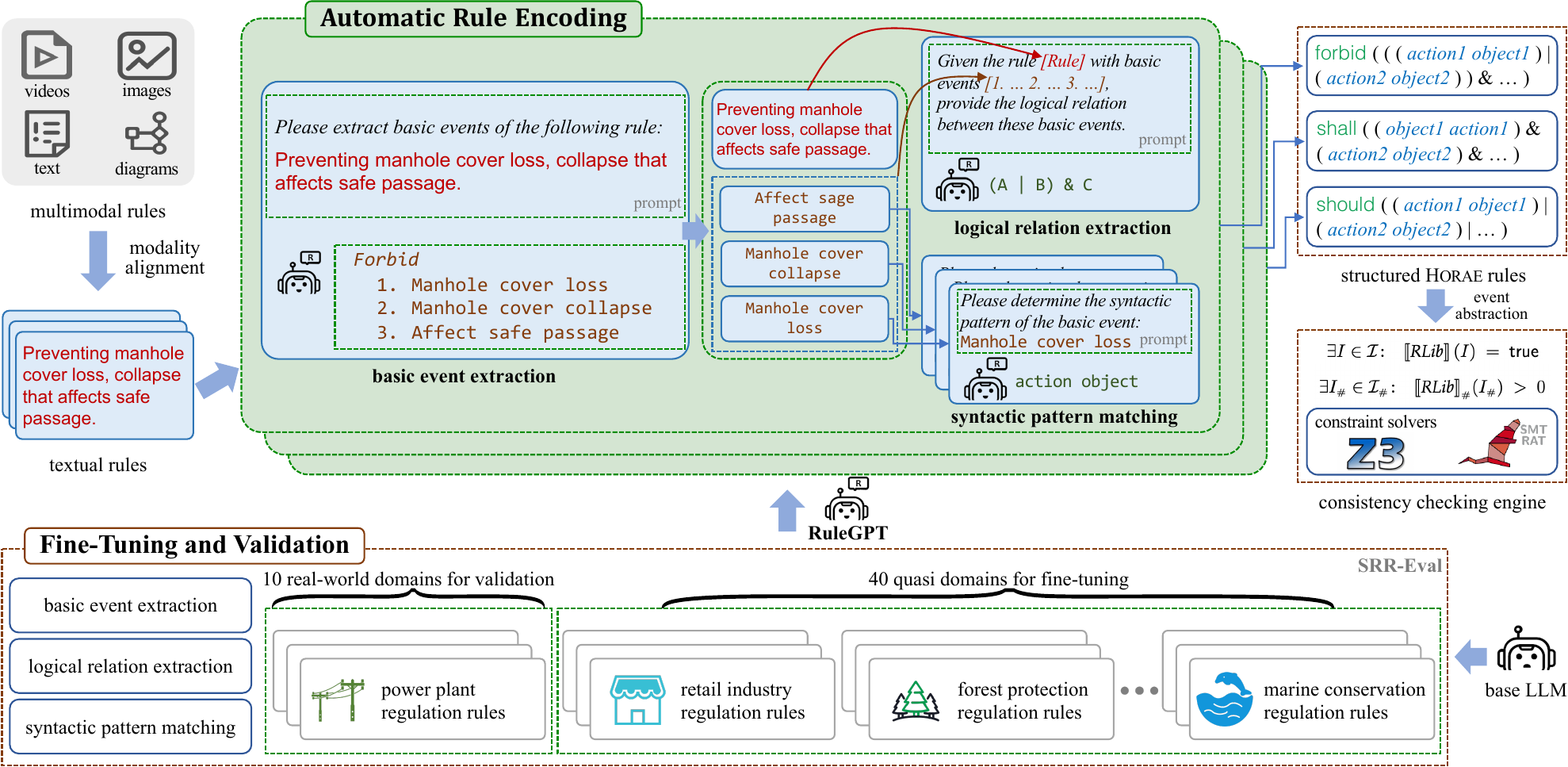}
	\caption{The overall process of automated transformation using the fined-tuned {\gptname}.}
	\label{rulegpt}
\end{figure*}

This section presents the fine-tuning process of {\gptname}. As key to automation in intelligent service regulation, {\gptname} aims to automatically convert regulation rules written in natural languages to their unified, structured {\langname} representations in the form of \emph{token streams} as depicted in 
\cref{rulegpt}. 

We note that off-the-shelf LLMs are not suitable for the above conversion task. The reasons are three-fold: 
\begin{enumerate*}[label=(\roman*)]
    \item existing LLMs are \emph{unaware} of {\langname} since its knowledge is not part of the corpus used to pre-train these models;
    \item closed-source, proprietary models like GPT-4o are prone to security issues as many regulation tasks are \emph{privacy-sensitive}; and
    \item general purpose LLMs, e.g., DeepSeek-R1~\cite{guo2025deepseek} and GPT-4o, require \emph{significant computational resources}. Moreover, they often exhibit low accuracies (see \cref{sec:experiments}) when performing the transformation in a \emph{monolithic} manner: Given a rule in natural language with a designed prompt, a general LLM cannot fully comprehend the basic events, logical relations, and syntactic patterns simultaneously and convert the rule into {\langname} under zero- or few-shot conditions.
\end{enumerate*}
To address these challenges, we propose to 
\begin{enumerate*}[label=(\roman*)]
    \item \emph{create} a benchmark dataset for service regulation rules ({\datasetname}, for short);
	\item \emph{fine-tune} a pre-trained, open-sourced LLM using {\datasetname} to encode the {\langname} knowledge; and
	\item \emph{partition the fine-tuning process} into three cooperative phases, i.e., basic event extraction, logical relation extraction, and syntactic pattern matching.
\end{enumerate*}


\paragraph*{\it Overview of \textnormal{\datasetnameNaked}}
{\datasetname} consists of 10 domains with real-world regulation rules (50 rules for each domain) and 40 domains with LLM-generated quasi rules (115 rules for each domain), amounting to \emph{50 domains with 5,100 rules} in total. 
SRR-Eval is open-sourced at \url{https://huggingface.co/datasets/Xfgll/SRR-Eval}. See Details in \aref{appx:benchmark}.

\paragraph*{\it Fine-Tuning Strategy}
We use LoRA (\emph{low-rank adaptation} \cite{hu2022lora}) to fine-tune our base model $M$. Let $W \in \Reals^{d \times q}$ be the pre-trained weight matrix of $M$. In contrast to full fine-tuning where all model parameters are retrained by augmenting $W$ with its accumulated gradient update $\Delta W \in \Reals^{d \times q}$, LoRA freezes $M$ and injects low-rank decomposition matrices $A \in \Reals^{d \times r}$ and $B \in \Reals^{r \times q}$ with trainable parameters into each layer of the transformer architecture, i.e.,
\begin{align*}
    W' \eeq W + A B~,
\end{align*}%
where $r \ll \min(d,q)$ is the rank of a LoRA module; $W' \in \Reals^{d \times q}$ is the adapted weight matrix. LoRA thus significantly reduces the number of trainable parameters.
%
We denote by
\begin{align*}
    M' \eeq \text{LoRA}\left(M,D\right)
\end{align*}%
the process of fine-tuning $M$ via LoRA into an adapted model $M'$ which incorporates the knowledge encoded in dataset $D$. 

\subsection{Extracting Basic Events}\label{sec:basiceventextration}

Given a textual regulation rule $R$ written in a natural language, the phase of \emph{basic event extraction} aims to fine-tune a pre-trained base LLM $M$ into a dedicated model $M_\textit{event}$ for extracting the set $E$ of basic events from $R$, i.e.,
\begin{align*}
    M_\textit{event}\colon \quad R \mmapsto \{e_1,e_2,\dots,e_m\} \ddefeq E~,
\end{align*}%
where every basic event $e_i$ is of a certain pattern adhering to the {\langname} syntax. See the composite rule \cref{eq:quasi-rule-poor-diversity} and its followed {\langname} form as an example of the extraction. Note that recognizing the specific event patterns is the task of the syntactic pattern matching phase as discussed in \cref{sec:syntacticpatternmatching}.

We obtain $M_\textit{event}$ by fine-tuning $M$ via LoRA, namely,
\begin{align*}
M_\textit{event} \eeq \text{LoRA}\left(M,D_\textit{event}\right)~,
\end{align*}%
i.e., we feed LoRA with a dedicated training dataset $D_\textit{event}$ sourced from {\datasetname}, which is formatted as
\begin{align*}
D_\textit{event} \eeq \left\{(u_i,a_i)\right\}^n_{i=1}
\end{align*}%
with $u_i$ being the \emph{user prompt} and $a_i$ the corresponding \emph{assistant's extraction}. Specifically, every entry $(u_i,a_i)$ in $D_\textit{event}$ is of the following query-response format:
\begin{equation*}
	\begin{aligned}
		u_i & \eeq \text{``}\rulefont{Please~extract~basic~events~of~the~following~rule\colon} \\[-1mm]
            &\quad\quad\,\;\rulefont{[original~rule]} \text{''} \\
		a_i & \eeq \text{``}\rulefont{[basic~events]} \text{''}
	\end{aligned}
\end{equation*}%
where $\rulefont{[original~rule]}$ and $\rulefont{[basic~events]}$ are raw ingredients of the \emph{composite quasi rules} in {\datasetname} (cf.\ \aref{subsec:dataset-construction}).

\subsection{Extracting the Logical Relation}\label{logicalrelationextraction}
In the phase of \emph{logical relation extraction}, we fine-tune a base LLM $M$ into a tailored model $M_\textit{logic}\colon (R, E) \mmapsto L$ for extracting the logical relation $L$ between basic events $E$ of $R$;
e.g., the logical relation of rule \cref{eq:quasi-rule-poor-diversity} is $L = \eventPlain_{11} \vee \eventPlain_{12} \vee \eventPlain_{13}$. We remark that the quality of the {\langname} transformation depends heavily on $M_\textit{logic}$ because logical relations are the key contributor in both the qualitative and the quantitative semantics of {\langname} as demonstrated in \cref{sec:semantics}.

Akin to the event extraction phase, $M_\textit{logic}$ is derived by
$M_\textit{logic} = \text{LoRA}(M,D_\textit{logic})$.
Here, the training dataset
$D_\textit{logic} = \{(u'_i,a'_i)\}^n_{i=1}$
consists of query-response pairs: 
\begin{equation*}
	\begin{aligned}
		u'_i & \eeq \text{``}\rulefont{Given~the~rule~[original~rule]~with~basic~events} \\[-1mm]
		&\quad\quad\,\;\rulefont{[basic~events],~provide~the~logical~relation}\\[-1mm]
		&\quad\quad\,\;\rulefont{between~these~basic~events} \text{''} \\
		a'_i & \eeq \text{``}\rulefont{[logical~relation]} \text{''}
	\end{aligned}
\end{equation*}%
where $\rulefont{[original~rule]}$, $\rulefont{[basic~events]}$, and $\rulefont{[logical~relation]}$ are raw data of \emph{composite quasi rules} in {\datasetname} (\cref{subsec:dataset-construction}). 

\subsection{Matching Syntactic Patterns}\label{sec:syntacticpatternmatching}

Let $T = \{t_1, t_2, \ldots, t_j\}$ be the fixed \emph{finite} set of syntactic patterns as defined in the bottom-level grammar of {\langname} in \cref{subsec:syntax}. The goal of \emph{syntactic pattern matching} is to attach to every basic event in $E$ a corresponding syntactic pattern in $T$ via a fine-tuned model 
$M_\textit{syntax}\colon E \to T$.

In analogous to $M_\textit{event}$ and $M_\textit{logic}$, $M_\textit{syntax}$ is obtained by
$M_\textit{syntax} = \text{LoRA}(M,D_\textit{syntax})$,
where the dedicated training dataset
$D_\textit{syntax} = \{(u''_i,a''_i)\}^n_{i=1}$
is composed of
\begin{equation*}
	\begin{aligned}
		u''_i & \eeq \text{``}\rulefont{Please~determine~the~syntactic~pattern~of~the} \\[-1mm]
		&\quad\quad\,\;\rulefont{basic~event\colon [basic~event]} \text{''} \\
		a''_i & \eeq \text{``}\rulefont{[syntactic~pattern]} \text{''}
	\end{aligned}
\end{equation*}%
where $\rulefont{[basic~event]}$ and $\rulefont{[syntactic~pattern]}$ are raw ingredients of the \emph{single-event quasi rules} in {\datasetname} (\aref{subsec:dataset-construction}). These ingredients are utilized to train {\gptname}'s ability to classify basic events $E$ into the set $T$ of syntactic categories.

By combining the aforementioned fine-tuned models, we obtain {\gptname} (see the general pipeline in \cref{rulegpt}):
\begin{align*}
	\gptname \eeq \!\left\{M_\textit{event},M_\textit{logic},M_\textit{syntax}\right\}~.
\end{align*}%


%
\section{Experimental Results}\label{sec:experiments}

This section presents an empirical evaluation of {\gptname}'s performance against several baselines. Our primary goal 
is to \emph{demonstrate the feasibility and effectiveness of {\gptname} in automating the modeling process in {\langname} across different real-world regulation domains}, which essentially enables our end-to-end framework for fully automated intelligent service regulation. 
{\gptname} is open-sourced \ifanonymous (Link for review: \url{https://anonymous.4open.science/r/Horae-6B73})\else via GitHub at \url{https://github.com/FICTION-ZJU/RuleGPT}\fi. 


\paragraph*{\it Settings of Fine-Tuning}
We implement {\gptname} by adapting -- via the LoRA technique~\cite{hu2022lora} -- Qwen2.5-7B-Ins~\cite{DBLP:journals/corr/abs-2412-15115} as our common base model shared by the three fine-tuning phases. The fine-tuning procedure is conducted on a single NVIDIA A100-40GB GPU. We set the learning rate to $1 \times 10^{-4}$ and employ gradient accumulation with 16 steps to effectively manage the computational load. The training spans 3 epochs, we use bf16 precision to assist in managing GPU memory efficiently and employ gradient checkpointing to further optimize the memory usage. The fine-tuning datasets are sourced from {\datasetname} as described in \aref{appx:benchmark}; a set of hyperparameters, e.g., weight decay (0.1), Adam optimizer's $\beta_2$ (0.95), warmup ratio (0.01), and cosine learning rate scheduler (enable) further contributes to the training stability and efficiency.


\paragraph*{\it Baselines}
We compare {\gptname} against three baselines: Qwen2.5-7B-Ins, GPT-3.5(-Turbo), and GPT-4o(-latest). The latter two, though being closed-source models, are chosen because
\begin{enumerate*}[label=(\roman*)]
	\item they are widely recognized for their capabilities in natural language understanding and generation; and
        \item models with 7B parameters may outperform GPT-3.5 in certain scenarios, as observed in~\cite[Sect.~3.3]{DBLP:journals/corr/abs-2309-16609}.
\end{enumerate*}

In the rest of this section, we present detailed experimental results with respect to the three fine-tuning phases.

\subsection{Basic Event Extraction}\label{sec:ex-event-extraction}

For the component of basic event extraction, we compare {\gptname} against the baselines in terms of three performance metrics: the \emph{precision} $\precision$, the \emph{recall} $\recall$, and the \emph{$F_1$-score} $\fscore$ (i.e., the harmonic mean of $\precision$ and  $\recall$). These metrics together provide a comprehensive assessment of the models' accuracy and adaptability in extracting basic events in {\langname}. The details of these metrics are presented in \aref{appx: evaluation metrics}

\begin{figure}[t]
\centering
\begin{subfigure}[b]{0.32\linewidth}
    \centering
    \includegraphics[width=1\linewidth]{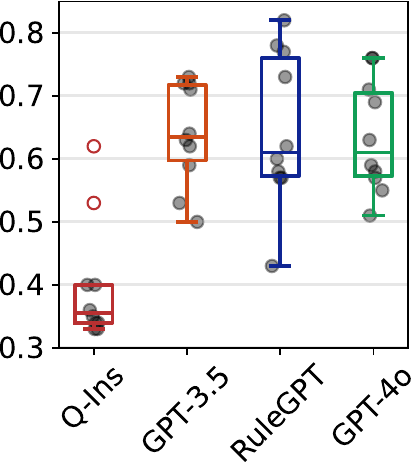}
    \caption{The precision $\precision$}
    \label{fig:boxplot-P}
\end{subfigure}
\hfil
\begin{subfigure}[b]{0.32\linewidth}
    \centering
    \includegraphics[width=1\linewidth]{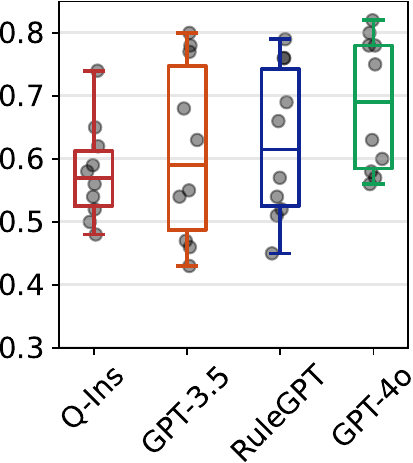}
    \caption{The recall $\recall$}
    \label{fig:boxplot-R}
\end{subfigure}
\hfil
\begin{subfigure}[b]{0.32\linewidth}
    \centering
    \includegraphics[width=1\linewidth]{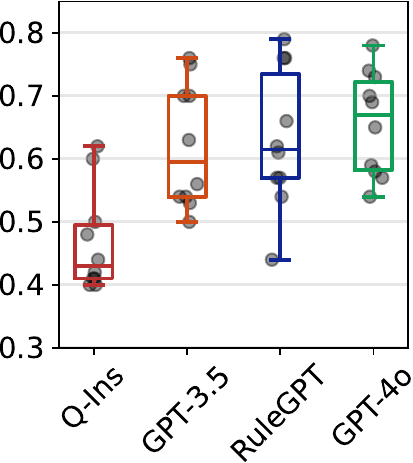}
    \caption{The $F_1$-score $\fscore$}
    \label{fig:boxplot-F1}
\end{subfigure}
\caption{Visualization of data in \cref{tab:experiment-results-event} (Q-Ins abbreviates Qwen2.5-7B-Ins). 
Every scattered boxplot 
depicts the corresponding 
column of \cref{tab:experiment-results-event} 
with its five-number summary.
}
\label{fig:boxplots}
\end{figure}

We report our experimental results w.r.t.\ basic event extraction in \cref{tab:experiment-results-event}, where we mark the \best{best} results and the \underline{second-best} results among all the competitors. The scattered boxplots in \cref{fig:boxplots} further visualize these numerical results separately for the three metrics. The following observations are drawn from these results:
\begin{enumerate*}[label=(\roman*)]
    \item {\gptname} significantly outperforms its base model Qwen2.5-7B-Ins in all three metrics, thus demonstrating the feasibility and effectiveness of our fine-tuning process and the quality of SRR-Eval.
    \item For the precision metric, {\gptname} is the winner amongst all the models -- it achieves the best results over 6/10 benchmarks.
    \item For the recall metric, {\gptname} exhibits a comparable ability with GPT-3.5, but they both are slightly inferior to GPT-4o. 
    \item For the $F_1$-score metric, \emph{{\gptname} performs better than \textnormal{GPT-3.5}}, \emph{slightly inferior to \textnormal{GPT-4o}}. 
\end{enumerate*}

\subsection{Logical Relation Extraction}\label{sec:ex-logic-extraction}

Next, we compare {\gptname} against the baselines in terms of the \emph{accuracy} in extracting logical relations between basic events. Since the formal semantics of a {\langname} rule depends heavily on the underlying logical relation (see \cref{sec:semantics}), an extraction is considered \emph{correct} iff the extracted logical relation \emph{semantically coincides with} the relation in {\datasetname}.

\begin{table}[t]
\centering
\caption{Experimental results w.r.t.\ basic event extraction ($\precision$ for precision, $\recall$ for recall, and $\fscore$ for $F_1$-score).}
\label{tab:experiment-results-event}
\centering
\tiny
\setlength{\tabcolsep}{3pt}
\resizebox{\linewidth}{!}{%
\begin{tabular}{lcccccccccccc}
\toprule
\multirow{2}{*}{\makecell[l]{\vspace*{-2.2mm}\\\textbf{Real-world dataset}\\\textbf{in {\datasetnameNaked}}}}& \multicolumn{3}{c}{Qwen2.5-7B-Ins}  & \multicolumn{3}{c}{GPT-3.5}  & \multicolumn{3}{c}{\gptname} & \multicolumn{3}{c}{GPT-4o} \\
\cmidrule(l{3pt}r{3pt}){2-4}
\cmidrule(l{3pt}r{3pt}){5-7}
\cmidrule(l{3pt}r{3pt}){8-10}
\cmidrule(l{3pt}r{3pt}){11-13}
& $\precision$ & $\recall$ & $\fscore$  & $\precision$ & $\recall$ & $\fscore$ & $\precision$ & $\recall$ & $\fscore$ & $\precision$ & $\recall$ & $\fscore$  \\
\cmidrule(l{3pt}r{3pt}){1-1}
\cmidrule(l{3pt}r{3pt}){2-4}
\cmidrule(l{3pt}r{3pt}){5-7}
\cmidrule(l{3pt}r{3pt}){8-10}
\cmidrule(l{3pt}r{3pt}){11-13}
power plant &0.40  &0.58  &0.48   &0.50 &0.63  &0.56  &\underline{0.62}  &\underline{0.69}  &\underline{0.66} &\best{0.71} &\best{0.78} &\best{0.74}   \\
public place safety &0.40  &0.65&0.50  &0.72  &\underline{0.80}  &\underline{0.76}  &\best{0.77}  &0.76  &\underline{0.76} &\underline{0.76} &\best{0.82} &\best{0.78}    \\
tourism &0.34  &0.62  &0.44    &\underline{0.71}  &\best{0.78} &\underline{0.75}  &\best{0.82} &\underline{0.76} &\best{0.79}  &0.69 &\best{0.78} &0.73 \\
energy regulation &0.62  &0.59  &0.60    &0.73  &\underline{0.55}  &\underline{0.63}  &\best{0.78}  &0.51 &0.62 &\underline{0.76} &\best{0.63} &\best{0.69}  \\
urban management &0.53  &0.74  &0.62    &\underline{0.64}  &0.77 &\underline{0.70} &\best{0.73} &\underline{0.79} &\best{0.76} &0.63  &\best{0.80}  &\underline{0.70} \\
forest products &0.35  &0.48  &0.40    &\best{0.63}  &0.47  &\underline{0.54}  &\underline{0.57} &\underline{0.52}  &\underline{0.54} &0.55  &\best{0.60}  &\best{0.57}  \\
tabacco &0.33  &0.56  &0.41    &\best{0.72}  &\underline{0.68}  &\best{0.70}  &\underline{0.58}  &0.66  &0.61  &0.57  &\best{0.75}  &\underline{0.65} \\
agricultural markets &0.34  &0.50  &0.40    &0.59  &0.43  &0.50  &\best{0.60}  &\underline{0.54}  &\underline{0.57}  &\underline{0.58}  &\best{0.57}  &\best{0.58}  \\
food safety &0.33  &0.54  &0.41    &\underline{0.53}  &0.54  &\underline{0.54}  &\best{0.57} &\best{0.57}  &\best{0.57} &0.51  &\underline{0.56}  &\underline{0.54}    \\
forest degradation &0.36  &0.52  & 0.42   &\best{0.62}  &\underline{0.46}  &\underline{0.53}  &0.43  &0.45  &0.44 &\underline{0.59}  &\best{0.58}  &\best{0.59} \\
\bottomrule
\end{tabular}
}
\end{table}



The evaluation results w.r.t.\ logical relation extraction are reported in (the left part of) \cref{tab:experiment-results-relation-syntax}. It shows that \emph{{\gptname} exhibits the highest accuracy on par with \textnormal{GPT-4o} consistently over all the ten benchmarks}. More concretely, we make the following observations:
\begin{enumerate*}[label=(\roman*)]
    \item As the underlying base model of {\gptname}, Qwen2.5-7B-Ins performs poorly in identifying logical relations.
    \item However, our fine-tuning procedure suffices to optimize this small model 
    to perform better than the GPT-3.5, yielding a cost-effective and computationally efficient solution.
    \item The comparable performance of {\gptname} against GPT-4o indicates that, in our case, a small model fine-tuned with SRR-Eval can potentially replace larger proprietary models that are generally more resource-intensive.
\end{enumerate*}

\begin{table}[t]
      \centering
        \caption{Accuracy of logical relation extraction and syntactic pattern matching (Q-Ins is shorthand for Qwen2.5-7B-Ins).}
        \label{tab:experiment-results-relation-syntax}
        \centering\scriptsize
        \setlength{\tabcolsep}{1pt}
        \resizebox{.886\linewidth}{!}{%
        \begin{tabular}{lcccccccc}
            \toprule
            \multirow{2}{*}{\makecell[l]{\vspace*{-2.2mm}\\\textbf{Real-world dataset}\\\textbf{in {\datasetnameNaked}}}} & \multicolumn{4}{c}{\scriptsize{\textbf{Logical relation extraction}}} & \multicolumn{4}{c}{\scriptsize{\textbf{Syntactic pattern matching}}} \\
            \cmidrule(l{2pt}r{2pt}){2-5} \cmidrule(l{2pt}r{2pt}){6-9}
            & Q-Ins & GPT-3.5 & GPT-4o & {\gptname} & Q-Ins & GPT-3.5 & GPT-4o & {\gptname} \\
            \cmidrule(l{1pt}r{1pt}){1-1} \cmidrule(l{2pt}r{2pt}){2-5} \cmidrule(l{2pt}r{2pt}){6-9}
            power plant & 0.34 & 0.38 & \best{0.70} & 0.66 & 0.22 & 0.62 & 0.66 & \best{0.72} \\
            public place safety & 0.39 & 0.57 & 0.78 & \best{0.84} & 0.08 & 0.13 & \best{0.36} & 0.23\\
            tourism & 0.24 & 0.40 & 0.74 & \best{0.76} & 0.14 & 0.17 & 0.16 & \best{0.24}\\
            energy regulation & 0.11 & 0.24 & 0.4 & \best{0.73} & 0.06 & 0.23 & \best{0.65} & 0.39\\
            urban management & 0.22 & 0.38 & \best{0.80} & 0.60 & 0.11  & 0.17 & \best{0.40} & 0.26 \\
            forest products & 0.10 & 0.34 & \best{0.66} & 0.46 & 0.08 & 0.19 & \best{0.43} & 0.39\\
            tabacco & 0.14 & 0.36 & 0.58 & \best{0.66} & 0.18 & 0.17 & 0.29 & \best{0.47} \\
            agricultural markets & 0.10 & 0.34 & \best{0.60} & 0.52 & 0.36 & 0.08 & 0.24  & \best{0.44}\\
            food safety & 0.18 & 0.48 & 0.62 & \best{0.64} & 0.31 & 0.15 & \best{0.36} & 0.27\\
            forest degradation & 0.10 & 0.16 & \best{0.52} & 0.44 & 0.23 & 0.17 & 0.26 & \best{0.27}\\
            \midrule
            \footnotesize{Mean} &0.19   &0.37  &0.64   &0.63         &0.18 & 0.21 &0.38  &0.37 \\
            \footnotesize{Variance} &0.01   &0.11  &0.01  &0.01  &0.01 & 0.14  &0.02  &0.02  \\
            \bottomrule
        \end{tabular}
        }
\end{table}

\subsection{Syntactic Pattern Matching}\label{sec:ex-syntax-matching}

Finally, we compare {\gptname} against the baselines in terms of the \emph{accuracy} in matching syntactic patterns of basic events. 
As it is essentially a classification task, the result is considered \emph{correct} iff \emph{the correct syntactic category is identified}.

The experimental results w.r.t.\ syntactic pattern matching are reported in (the right part of) \cref{tab:experiment-results-relation-syntax}. We observe that \emph{{\gptname} achieves the highest accuracy over 5/10 benchmarks, which significantly outperforms \textnormal{Qwen2.5-7B-Ins} and the proprietary model \textnormal{GPT-3.5}}, and is on par with \textnormal{GPT-4o}. 
%


\paragraph*{\bf Overall Performance}
Our experiments demonstrate the overall feasibility and effectiveness of {\gptname} in automating the modeling process in {\langname} across different real-world regulation domains:
\begin{enumerate*}[label=(\roman*)]
    \item {\gptname} significantly outperforms \textnormal{GPT-3.5} in extracting logical relations and syntactic patterns, and performs on par with it in the task of basic event extraction.
    \item The substantial improvement of {\gptname} over Qwen2.5-7B-Ins underscores the effectiveness of our fine-tuning strategy, further demonstrating the high quality of \datasetname we have created.
    \item We show the feasibility of automating a complex task (i.e., {\langname} modeling) by breaking it down into simpler components (i.e., the three fine-tuned models), each of which is optimized individually and contributes to a highly effective overall system (i.e., {\gptname}).
\end{enumerate*}
%


%
\section{Related Work}\label{sec:related-work}

Service regulation strives to represent regulatory compliance requirements 
with modeling languages for automation \cite{DBLP:journals/is/MuehlenI10}:
The language SWRL \cite{horrocks2004swrl} enables complex reasoning in semantic web applications. BPMN-Q \cite{awad2011visually} visually specifies compliance rules and explains violations in business processes using a pattern-based approach to link BPMN-Q graphs with formal temporal logic expressions. CRL \cite{DBLP:journals/sosym/ElgammalTHP16} 
offers a comprehensive framework for managing business process compliance, which introduces abstract pattern-based specifications while supporting compensations and non-monotonic requirements. DecSerFlow \cite{DBLP:conf/wsfm/AalstP06} is a declarative 
language for specifying, enacting, and monitoring service flows, grounded in temporal logic to address the autonomous nature of services. An orthogonal line of research aims to evaluate the expressiveness and complexity of 
rule languages by leveraging real-world examples and normative classification frameworks, addressing the challenge of representing complex constraints across multiple process perspectives \cite{zasada2023evaluation}.

Our work is closely related to the rule language CDSRL and the LLM-based converter RegGPT recently proposed in \cite{DBLP:conf/icws/WangXZMKZ24} to model cross-domain regulatory requirements. 
The key differences are
\begin{enumerate*}[label=(\roman*)]
    \item {\langname} supports \emph{behavioral compositionality} by maintaining an abstracted layer of fine-grained basic events, thus admitting domain-agnostic downstream recognition models to discharge the regulation tasks.
    In contrast, CDSRL emphasizes holistic rule structuring without explicit behavioral decomposition;
    \item {\langname} admits \emph{formal semantics} that enable automated consistency checking and violation quantification through SMT solvers, whereas CDSRL lacks executable validation mechanisms beyond syntactic template matching;
    \item {\gptname} supports \emph{fully autonomous rule conversion} through phased fine-tuning of open-sourced models while RegGPT's conversion pipeline depends critically on GPT-4 and prompt templates. 
\end{enumerate*}

\section{Conclusion}\label{sec:conclusion}

We presented the domain-agnostic modeling language \langname. It enables an end-to-end intelligent regulation framework leveraging 
a fine-tuned LLM \gptname to automate the conversion of natural language regulation rules into a structured intermediate representation. {\langname} is, to the best of our knowledge, the first modeling language that admits \emph{fully automated} service regulation with effective domain-modality unification.
Future work includes integrating \langname and \gptname with downstream recognition models and algorithms to detect (quantitative) service-rule violations.







\ifanonymous
\else

\section*{Acknowledgments}
This work was partially supported by the National Key R\&D Program of China (No.\ 2022YFF0902600), by the ZJNSF Major Program (No.\ LD24F020013 and LD24F020014), by the Fundamental Research Funds for the Central Universities of China (No.\ 226-2024-00140), by the Zhejiang Pioneer Project (No.\ 2023C01G1752957), and by the ZJU Education Foundation's Qizhen Talent program. The authors would like to thank Linyu Yang for the helpful discussion on the formal semantics of {\langname}.
\fi

\bibliographystyle{named}
\bibliography{references}



\begin{appendices}

\section{Multimodal Rule Alignment}\label{subsec:Multi-Modality}

\begin{figure}[t]
	\centering
	\begin{tikzpicture}
		\draw (0, 0) node[inner sep=0] {\includegraphics[width=1\linewidth]{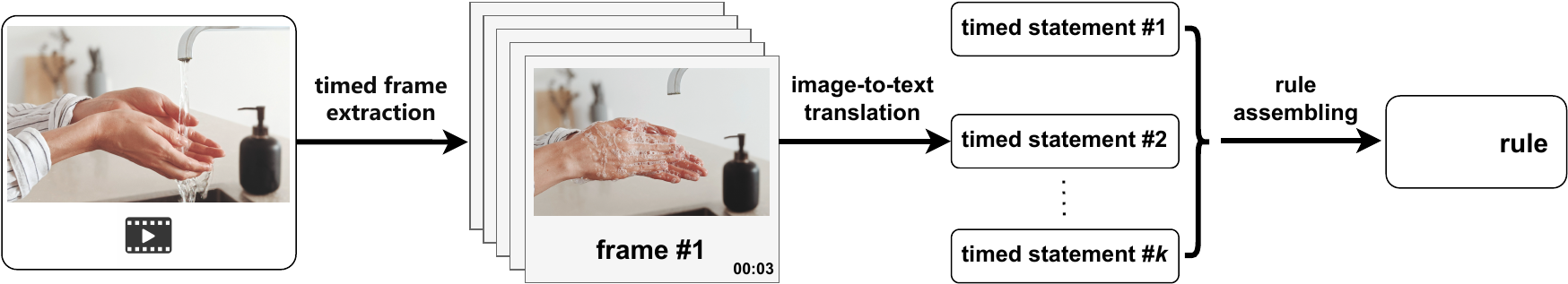}};
		\draw (3.76, -.016) node {\scaleto{\langname}{3pt}};
	\end{tikzpicture}
	\caption{An illustration of the video-to-text pipeline.}
	\label{fig:video_process}
\end{figure}

We show how multimodal regulation rules -- notably in the form of \emph{images} and \emph{videos} -- can be aligned to the text modality such that rules of different formats can be interpreted through a unified medium, i.e., {\langname} rules. Prominent examples of non-textual rules include a traffic sign which uses a simple image to convey the message of \enquote{no parking} and a videocast that effectively demonstrates the \enquote{seven-step handwashing method} (see \cref{fig:video_process}). Such multimodal rules often carry more rich and intuitive information than those in single textual mode. The former can be used either exclusively to describe the regulation intentions, or as a complement to the text modality to further illustrate and/or clarify the regulation rules.

However, aligning multimodal rules to the text modality is non-trivial: The primary challenge lies in the inherent disparity in representing information across different modalities and the intricacies induced by their integration. The structural gap between textual and pictorial information exemplifies this issue, as it requires advanced mechanisms to effectively combine them into a structured and unified rule. Moreover, the processing of multimodal data typically requires the deployment of advanced algorithms and models, which are further impelled to maintain the data's consistency and comprehensiveness.

In response to the above challenge, a straightforward solution is to \emph{transform multimodal regulation rules into latent embedding vectors}~\cite{radford2021learning} using advanced deep learning models, especially those based on a unified transformer architecture~\cite{vaswani2017attention}. Such transformation involves complex feature extraction and information encoding, aiming to capture the core semantic elements across different modalities through a uniform embedding representation. However, due to the black-box nature of deep learning models, the resulting embedding vectors can be difficult to interpret in a structured manner. Furthermore, the performance of these models depends highly on the quality and scope of the pre-training data; this may significantly limit their range of applications.

Therefore, we propose to \emph{transform multimodal regulation rules into the text modality} on which {\langname} is based. Such transformation unleashes multiple features of {\langname}, e.g., structured interpretation and consistency checking, for their general applicability to regulation rules in non-textual modalities. Below, we showcase the video-to-text transformation, which in principle subsumes the image-to-text case. \Cref{fig:video_process} depicts our pipeline for converting a video (demonstrating the seven-step handwashing method -- a common rule in health service) to a textual rule in {\langname}. The pipeline consists of the following main steps:
\begin{enumerate}[itemsep=.8mm]
	\item \emph{Frame Extraction}: We extract a set of frames from the video in the form of a temporally contiguous sequence of images, i.e., timestamped snapshots of the original regulation rule. We exploit two extraction strategies: uniform frame extraction (i.e., sampling frames uniformly along the time axis) and key frame extraction (i.e., extracting a number of representative key frames). The former is simple and ensures a comprehensive overview of the entire video but can include irrelevant or redundant frames; the latter focuses on the most informative parts of a video but may miss subtle yet relevant information as observed in~\cite{tang2023deep}. We observe in our experiments that uniform frame extraction exhibits already a promising performance in terms of representativeness and efficiency.
	
	\item \emph{Image-to-Text Translation}: The sequence of timestamped images is then translated into a corresponding sequence of timed {\langname} statements leveraging advanced multimodal models available today, including proprietary ones, e.g., GPT-4(V)~\cite{openai2024gpt4}, as well as open-sourced models such as Qwen-VL~\cite{Qwen-VL} and LLaVA-NeXT~\cite{liu2024llavanext}. To evaluate the performance of these models, we construct a multimodal rule dataset consisting of 100 image rules together with their corresponding textual rules. These image rules are carefully curated such that they can be accurately translated into the textual form. To attain diversity, the set of images covers abstract symbols, real-world scenes, comics, and AIGC models. Experimental results show that GPT-4V achieves an accuracy of 81\%, indicating its effectiveness in image-to-text translation.
	
	\item \emph{Rule Assembling}: The sequence of timed {\langname} statements can be straightforwardly assembled by a general LLM -- in accordance with their chronological relations -- into a bulk of text in natural language, which can then be automatically encoded as a {\langname} rule by {\gptname}.
\end{enumerate}

\begin{figure*}[t]
	\begin{align*}
		s_1 &\eeq \underbrace{\makecell{\rulefont{employee~requests}\\\rulefont{an~annual~leave}}}_{\mathclap{\event_{\textcolor{NavyBlue}{11}}}}~\bigwedge~\bigg( \underbrace{\makecell{\rulefont{it~is~during~the}\\\rulefont{permissible~period}}}_{\mathclap{\event_{\textcolor{NavyBlue}{12}}}}~\bigvee~\underbrace{\makecell{\rulefont{manager~approves}\\\rulefont{the~request}}}_{\mathclap{\event_{\textcolor{NavyBlue}{13}}}} \bigg)~\longrightarrow~\underbrace{\rulefont{leave~is~granted}}_{\mathclap{\event_{\textcolor{NavyBlue}{14}}}}\\
		s_2 &\eeq \underbrace{\makecell{\rulefont{there~remains~sufficient}\\\rulefont{leave~balance}}}_{\mathclap{\event_{\textcolor{NavyBlue}{21}}}}~\bigwedge~\underbrace{\makecell{\rulefont{a~staff~member~files}\\\rulefont{for~vacation~days}}}_{\mathclap{\event_{\textcolor{NavyBlue}{22}}}}~\bigwedge~\underbrace{\makecell{\rulefont{manager~denies}\\\rulefont{the~request}}}_{\mathclap{\event_{\textcolor{NavyBlue}{23}}}}~\longrightarrow~\underbrace{\rulefont{administrative~review~is~required}}_{\mathclap{\event_{\textcolor{NavyBlue}{24}}}}\\
		s_3 &\eeq \bigg(\underbrace{\makecell{\rulefont{leave~balance~for~the}\\\rulefont{year~is~depleted}}}_{\mathclap{\event_{\textcolor{NavyBlue}{31}}}}~\bigvee~\underbrace{\makecell{\rulefont{a~worker~applies~for}\\\rulefont{yearly~leave~entitlement}}}_{\mathclap{\event_{\textcolor{NavyBlue}{32}}}}\bigg)~\bigwedge~\underbrace{\makecell{\rulefont{application~is~within}\\\rulefont{a~restricted~period}}}_{\mathclap{\event_{\textcolor{NavyBlue}{33}}}}~\longrightarrow~\underbrace{\rulefont{request~is~automatically~declined}}_{\mathclap{\event_{\textcolor{NavyBlue}{34}}}}
	\end{align*}%
	\caption{An illustration of event abstraction: SBERT suffices to detect (with appropriately chosen threshold) event correlations as $\event_{\textcolor{NavyBlue}{11}} = \event_{\textcolor{NavyBlue}{22}} = \event_{\textcolor{NavyBlue}{32}}$, $\event_{\textcolor{NavyBlue}{12}} = \neg \event_{\textcolor{NavyBlue}{33}}$, $\event_{\textcolor{NavyBlue}{13}} = \neg \event_{\textcolor{NavyBlue}{23}}$, $\event_{\textcolor{NavyBlue}{21}} = \neg \event_{\textcolor{NavyBlue}{31}}$, and $\event_{\textcolor{NavyBlue}{14}} = \neg \event_{\textcolor{NavyBlue}{34}}$.}\label{fig:event-abstraction}
\end{figure*}

\section{Design Principles}\label{appx:design-principles}
To achieve an effective modeling language for intelligent service regulation, the following principles are employed throughout the design of {\langname}:
\begin{itemize}[leftmargin=16pt,labelwidth=8pt,labelsep=6pt,itemsep=.8mm]
	\item \emph{Generality}: A domain-agnostic language shall accommodate regulation rules across different domains and/or multilingual texts, which may yield significant discrepancies in terms of, e.g., domain-specific terminologies, patterns, and writing styles. Hence, it requires a general paradigm that abstracts away domain-specific ingredients while still being able to express common regulation patterns. 
	\item \emph{Structuration}: Albeit with the generality principle, the language to be designed shall remain (semi-)structured such that
	\begin{enumerate*}[label=(\roman*)]
		\item regulation rules can be effectively stored, checked, and manipulated; and
		\item the potential ambiguity can be resolved as much as possible.
	\end{enumerate*}
	\item \emph{Automation}: The language shall feature simple patterns and structures, thereby allowing for an automated translation of regulation rules written in natural languages to those in the modeling language by means of, e.g., LLMs.
	\item \emph{Quantification}: Due to inherent impreciseness of downstream recognition models and algorithms, the language shall encode not only the qualitative information about a regulation rule -- i.e., whether it is satisfied or not -- but also the quantitative aspect of such satisfaction, i.e., the likelihood of it being satisfied.
\end{itemize}

\section{Event Abstraction}\label{appx:event-abstraction} 

Event abstraction refers to the process of transforming basic events -- which may be semantically correlated in natural language -- into symbolic propositions while preserving their semantic correlations. This can be achieved by various techniques established in natural language processing for assessing \emph{text similarity}, such as TF-IDF (term frequency-inverse document frequency~\cite{DBLP:journals/jd/Jones04}), Word2Vec~\cite{DBLP:journals/corr/abs-1301-3781}, and BERT (bidirectional encoder representations from transformers~\cite{DBLP:conf/naacl/DevlinCLT19}). Amongst the state-of-the-art methods, SBERT (Sentence-BERT~\cite{DBLP:conf/emnlp/ReimersG19}) has demonstrated the best performance. It utilizes a Siamese BERT architecture where two sentences are processed by two identical subnetworks (i.e., BERT models); the Siamese model is trained using contrastive learning to generate more accurate sentence embeddings. To further improve the accuracy, we apply a pre-trained SBERT model over the set of basic events -- instead of statements, where the logical relations are difficult for SBERT
to capture -- thereby yielding two types of correlations: $\eventPlain = \eventPlain'$ and $\eventPlain = \neg \eventPlain'$, as demonstrated by \cref{fig:event-abstraction}. Such an event-abstraction process facilitates the subsequent consistency check and may significantly reduce the number of event propositions therein.

\section{Benchmark}\label{appx:benchmark}

This part describes {\datasetname} -- our \emph{benchmark dataset for service regulation rules}. This dataset is primarily used for
\begin{enumerate*}[label=(\roman*)]
	\item extracting key observations guiding the design of {\langname} as demonstrated in \cref{sec:language};
	\item fine-tuning {\gptname} to automate the modeling process in {\langname}; and
	\item evaluating future cross-domain models geared toward intelligent service regulation.
\end{enumerate*}
{\datasetname} is open-sourced \ifanonymous (the link is omitted due to anonymous review)\else via Hugging Face at \url{https://huggingface.co/datasets/Xfgll/SRR-Eval}\fi.

The roadmap of constructing {\datasetname} is as follows: Albeit with numerous existing LLM datasets for testing general model understanding abilities, we need to \emph{create a dedicated dataset} for training and evaluating the model's understanding of regulation rules (written in natural language) and it's effectiveness to compile these rules into the {\langname} language. To this end, we start by \emph{collecting real-world regulation rules} with manually annotated {\langname} forms, which is, however, only possible for a few service regulation domains due to copyright and/or legal reasons. For instance, the ISO (International Organization for Standardization) standards encoding plenty of regulation rules are not freely distributed to the general public; financial services involving trade secrets typically do not allow direct access to their regulation rules. Therefore, we diversify our benchmark by \emph{generating quasi regulation rules} and their corresponding {\langname} forms leveraging state-of-the-art LLMs. Nonetheless, the so-obtained quasi rules may exhibit semantic contradictions against common sense and/or poor diversity in syntactic patterns, and hence we devise a method that combines conditional generation with manual screening to \emph{prune out semantic contradictions} and \emph{enforce syntactic diversity}.

\begin{figure*}[t]
	\centering
	\includegraphics[width=1\linewidth]{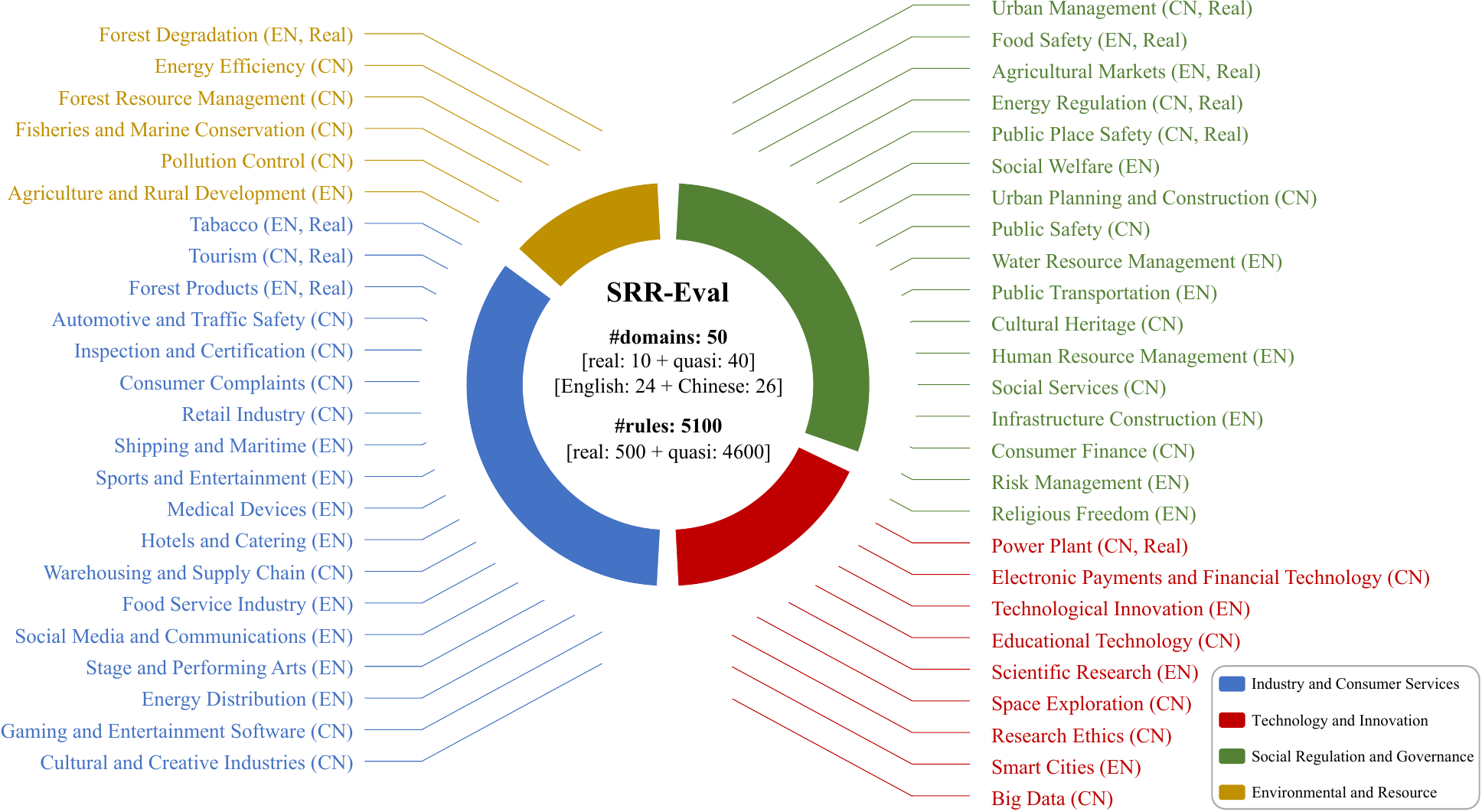}
	\caption{Overview of the bilingual {\datasetnameNaked} dataset.  \enquote{EN} and \enquote{CN} indicate rules in English and Chinese, respectively; \enquote{Real} marks domains where we collect real-world rules and unmarked ones are those with LLM-generated quasi rules.}
	\label{fig:dataset}
\end{figure*}

Below, we first give an overview of the {\datasetname} dataset, then provide technical details of collecting, generating, and formatting rules in {\datasetname}, and finally present a quality assessment of the underlying manual annotation process.

\paragraph*{\bf Overview of {\datasetnameNaked}}
\Cref{fig:dataset} sketches a bird's-eye perspective of our multilingual {\datasetname} dataset. It consists of 10 domains with real-world regulation rules (50 rules for each domain) and 40 domains with LLM-generated quasi rules (115 rules for each domain), thus amounting to \emph{50 domains with 5,100 rules} in total. All the involved domains are categorized into 4 fields of services covering industry and consumer services, technology and innovation, social regulation and governance, as well as environmental and resources. Every rule in {\datasetname} is structured and represented as a JSON object encoding its {\langname} form, which is obtained by either manual annotations (for real-world rules) or LLM-based generation (for quasi rules).

\subsection{Dataset Construction}\label{subsec:dataset-construction}

\paragraph*{\it Collection}
We employ the following principles when collecting real-world regulation rules:
\begin{itemize}
  \item Ensure significant \emph{heterogeneity} between different domains to better test the model's ability to understand rules in various domain contexts.
  \item Seek out \emph{multilingual} rule data to assess the model's adaptability to rules in multiple languages.
  \item Ensure that the collected data is free, public, and available for download and editing.
\end{itemize}
In line with these principles, we have collected a total of 500 real-world rules evenly distributed across 10 regulation domains (see \cref{fig:dataset}), as well as their manually annotated {\langname} forms. These real-world rules will later be used to \emph{validate} the performance of our fine-tuned {\gptname}.

\paragraph*{\it Generation}
We use GPT-4~\cite{openai2024gpt4} to generate quasi regulation rules covering 40 additional domains (cf.\ \cref{fig:dataset}) and their corresponding {\langname} forms. These quasi rules can be used either to \emph{train and test} cross-domain models like deep neural networks or, as in our case, to fine-tune existing LLMs. As an example, a quasi rule generated by GPT-4 for the domain of food safety may read as follows:
\begin{equation}\label{eq:quasi-rule-effective}
\begin{aligned}
	&\rulefont{Milk~cannot~be~sold~at~the~original~price} \\[-1mm]
	&\rulefont{within~\nblue{24}~hours~\green{before}~expiration.}
\end{aligned}\tag{R1}
\end{equation}%
Although Rule \cref{eq:quasi-rule-effective} deviates from its real-world counterpart, which may require a time window of $\rulefont{\nblue{12}~hours}$ instead of $\rulefont{\nblue{24}}$, it is considered to be \emph{effective} for training and testing the general ability to encode regulation rules in {\langname}. In contrast, the quasi rule
\begin{equation}\label{eq:quasi-rule-ineffective}
\begin{aligned}
	&\rulefont{Milk~cannot~be~sold~at~the~original~price} \\[-1mm]
	&\rulefont{within~\nblue{24}~hours~\maroon{after}~expiration.}
\end{aligned}\tag{R2}
\end{equation}%
is deemed \emph{ineffective} since it exhibits a \emph{semantic contradiction against common sense} and thus shall be excluded from the dataset. Such ineffective quasi rules are pruned out by means of a manual screening process, thereby yielding 50 effective quasi rules for each of the 40 domains.


We further observe that the LLM-generated quasi rules often exhibit \emph{poor diversity in syntactic patterns}: For instance, a typical quasi rule in the domain of big data reads as follows:
\begin{equation}\label{eq:quasi-rule-poor-diversity}
	\begin{aligned}
		&\rulefont{The~collected~information~should~include~user~behavior} \\[-1mm]
		&\rulefont{data,~user~preference~data,~or~user~transaction~data.}
	\end{aligned}\tag{R3}
\end{equation}%
The {\langname} form of this quasi rule is
\begin{align*}
	\should \left( \event_{\textcolor{NavyBlue}{11}} \vee \event_{\textcolor{NavyBlue}{12}} \vee \event_{\textcolor{NavyBlue}{13}} \right)\quad \text{with}\\[-1.5\baselineskip]
\end{align*}%
\begin{align*}
	&\event_{\textcolor{NavyBlue}{11}} = \rulefont{The~collected~information~include~user~behavior~data.}\\[-1mm]
	&\event_{\textcolor{NavyBlue}{12}} = \rulefont{The~collected~information~include~user~preference~data.} \\[-1mm]
	&\event_{\textcolor{NavyBlue}{13}} = \rulefont{The~collected~information~include~user~transaction~data.}
\end{align*}%
where all the basic events $\event_{\textcolor{NavyBlue}{11}}$, $\event_{\textcolor{NavyBlue}{12}}$, and $\event_{\textcolor{NavyBlue}{13}}$ are of the same syntactic pattern, i.e., $\obj~\act~\obj$. 
The LLM-generated quasi rules of the form \cref{eq:quasi-rule-poor-diversity} are appropriate to train the model's ability to deconstruct original rules and extract the underlying logical relation between basic events. However, due to the poor diversity in syntactic patterns, they are \emph{insufficient} to train the model's ability to classify basic events into specific syntactic categories (cf.\ the bottom-level grammar of {\langname} in \cref{subsec:syntax}).

To enforce syntactic diversity, we use GPT-4 to generate 65 additional quasi rules for each of the 40 domains. These rules do not feature complex logical relations; rather, they are single-event rules covering all event patterns in {\langname}. For example, to match the pattern $\obj\mydot\attr~\diamond~\val$ in the domain of taxi services, GPT-4 generates	
\begin{align}\label{eq:quasi-rule-single-event}
	\rulefont{The~response~delay~of~orders~shall~not~exceed~10mins.}\tag{R4}
\end{align}%

\paragraph*{\it Formatting}
Our {\datasetname} dataset is represented as a JSON array. Every rule in the dataset is structured and represented as a JSON object in the array encoding its {\langname} form, which is obtained by either manual annotations (for real-world rules, used for validation) or LLM-based generation (for quasi rules, used for training and testing).

A real-world rule in the \emph{validation set} is encoded as a JSON object with the following key-value pairs:
\begin{itemize}
\item \jstring{original rule}: string(rule), 
\item  \jstring{basic events}: list[string(event)], 
\item  \jstring{logical relation}: string(symbolic representation),
\item  \jstring{syntactic patterns}: list[string(pattern)].
\end{itemize}
Here, the list of basic events and the list of syntactic patterns are of the same length; the logical relation is captured by a string, e.g., \jstring{A \& B \& (C $\vert$ D)} for encoding $A \land B \land (C \lor D)$, where $A$, $B$, $C$, and $D$ correspond to the first, second, third, and fourth basic event in the list, respectively.

The quasi rules in the \emph{training and testing set} are encoded as JSON objects of two distinct forms. The first form encodes composite rules like \cref{eq:quasi-rule-poor-diversity} for training the ability to deconstruct original rules and extract the logical relation between basic events; it consists of three key-value pairs: 
\begin{itemize}
\item \jstring{original rule}: string(rule), 
\item  \jstring{basic events}: list[string(event)], 
\item  \jstring{logical relation}: string(symbolic representation).
\end{itemize}
We guide GPT-4 to generate these composite rules by specifying a variety of logical relations. Such guidance remedies the imbalance between logical connectives $\land$ and $\lor$.
The second JSON form encodes single-event rules like \cref{eq:quasi-rule-single-event} for training the ability to classify basic events into the given set of syntactic categories; it consists of two key-value pairs: 
 \begin{itemize}
\item  \jstring{basic events}: list[string(event)], 
\item  \jstring{syntactic patterns}: list[string(pattern)].
\end{itemize}
Akin to the first form, we can improve the syntactic diversity by enforcing syntactic patterns when generating the rules.

\subsection{Quality Assessment}\label{subsec:quality-assess}

To ensure the quality of manual annotations used for obtaining the {\langname} forms of real-world regulation rules, we provide training to volunteers on the task of annotation. We offer a concise tutorial and provide a representative set of example annotations as references. 
For the annotation process, the rules of each domain are annotated by three volunteers independently. If the annotated {\langname} forms are syntactically consistent among the volunteers, the corresponding form is taken as the final result; Otherwise, the principle of \enquote{majority rules} is adopted.

We employ \emph{Fleiss' kappa}~\cite{fleiss1971measuring} -- a statistical measure used to evaluate the reliability and agreement among multiple annotators when 
classifying data into multiple categories. For each of the 10 regulation domains with real-world rules, the corresponding value of Fleiss' kappa is 
\begin{align*}
	     \kappa \eeq \frac{P_o - P_e}{1 - P_e}~,
\end{align*}%
where $P_o$ is the \emph{observed agreement proportion} among annotators and $P_e$ is the \emph{expected agreement proportion}. Fleiss' kappa thus allows us to quantify the level of agreement beyond what would be expected by chance alone, thus providing valuable insights into the level of agreement among the annotators. For instance, the values of Fleiss' kappa for the domain of forest resource management and that of urban management are 0.845 and 0.864, respectively. These results indicate that our annotation consistency is at an \emph{almost perfect} agreement level ($0.81 \leq \kappa \leq 0.99$) as per \cite{landis1977measurement}.

\section{Performance Metrics}\label{appx: evaluation metrics}

To calculate the performance metrics, we first employ SBERT~\cite{DBLP:conf/emnlp/ReimersG19} to quantify the \emph{semantic similarity} between two basic events written in natural language:
\begin{align*}
	S(e_i,e_j) \ddefeq \cos\left(\textnormal{SBERT}(e_i),\textnormal{SBERT}(e_j)\right)~,
\end{align*}%
where $\textnormal{SBERT}(e)$ represents the specific vector encoding of event $e$ (see~\cite{DBLP:conf/emnlp/ReimersG19} for technical details of the encoding); $S(e_i,e_j) \in [0,1]$ denotes the cosine similarity between the two vectors encoding $e_i$ and $e_j$. Note that a larger value of $S(\cdot,\cdot)$ indicates a higher similarity.

We then define the \emph{precision} of basic event extraction as
\begin{align*}
	\precision \ddefeq \frac{1}{G} \cdot \sum_{i=1}^{G} S(e_i, e'_i)~,
\end{align*}%
where $G$ is the total number of basic events \emph{generated by the model} (for the whole rule library), $e_i$ is the $i$-th basic event generated by the model, and $e'_i$ is the \emph{original} basic event (provided in the dataset) of the same rule that is most similar to $e_i$. If such $e'_i$ does not exist, we simply set $S(e_i, \cdot)=0$.

Similarly, the \emph{recall} of basic event extraction is defined as
\begin{align*}
	\recall \ddefeq \frac{1}{O} \cdot \sum_{j=1}^{O} S(e'_j, e_j)~,
\end{align*}%
where $O$ is the total number of \emph{original} basic events in the dataset, $e'_j$ is the $j$-th original basic event, and $e_j$ is the \emph{generated} basic event of the same rule that is most similar to $e'_j$. If such $e_j$ does not exist, we set $S(e'_j, \cdot)=0$.
%
%

To symmetrically represent both precision and recall in one metric, the \emph{$F_1$-score} (of basic event extraction) takes the harmonic mean of precision and recall, i.e.,
\begin{align*}
	\fscore \ddefeq 2 \cdot \frac{\precision \cdot \recall}{\precision + \recall}~.
\end{align*}%

\end{appendices}

\end{document}